\documentclass{article}
\usepackage[utf8]{inputenc}
\usepackage{graphicx}
\usepackage{hyperref}
\usepackage{amsmath}
\usepackage{amssymb}
\usepackage{caption}
\usepackage{subcaption}
\usepackage{xcolor}
\usepackage{hyperref}
\usepackage{bm}
\usepackage{algorithm2e}
\usepackage{geometry}
\DeclareMathOperator{\sgn}{sgn}

\newcommand{\bb}[1]{\mathbf{#1}}

\title{Ten Years after ImageNet:\\ A $360^{\circ}$ Perspective on AI}
\author{Sanjay Chawla$^1$ \; Preslav Nakov$^2$
\; Ahmed Ali$^1$ 
\;  Wendy Hall$^3$ \\
Issa Khalil$^1$ \;
 Xiaosong Ma$^1$ \;
 Husrev Taha Sencar$^1$ \; \\
 Ingmar Weber$^5$ \;
Michael Woolridge$^4$ \;
Ting Yu$^1$ }
\date{
    $^1$Qatar Computing Research Institute\\%
    $^2$ Mohamed Bin Zayed University of AI \\
    $^3$University of Southampton \\
    $^4$ Oxford University\\
    $^5$ Saarland University \\[2ex]
    \today
}

\begin{document}

\maketitle

\begin{abstract}
It is ten years since neural networks made their spectacular comeback. Prompted by this anniversary, we take a holistic perspective on Artificial Intelligence (AI).
Supervised Learning for cognitive tasks is effectively solved --- provided we have enough high-quality labeled data. However, deep neural network models are not easily interpretable, and thus
the debate between blackbox and whitebox modeling has come to the fore. The rise of attention networks, self-supervised
learning, generative modeling, and graph neural networks has widened the application space of AI. Deep Learning has also propelled the return of reinforcement learning as a core building block of autonomous decision making systems. The possible harms made possible by new AI technologies have raised socio-technical issues such as transparency, fairness, and accountability. The dominance of AI by Big-Tech  who control talent, computing resources, and most importantly, data may lead to an extreme AI divide. Failure to meet high expectations in high profile, and much heralded flagship projects like self-driving vehicles could trigger another AI winter.
\end{abstract}

\newpage
\tableofcontents
\newpage
\section{Introduction}
The ImageNet challenge for automatically recognizing and labeling objects in images was launched
in 2010~\cite{imagenet_cvpr09}. However, it was in 2012 when AlexNet, an eight-layer (hence deep) convolutional neural network (CNN) emerged as the winner by a large
margin, and ushered in the new era of  AI~\cite{alexnet}. CNNs were not new and had been proposed as far back as the 1990s,
but had been sidelined in favor of more theoretically rigorous ML approaches such as support vector machines (SVMs) and
boosting methods~\cite{lecun1995convolutional,freund1997decision,scholkopf2002learning}. So, why did CNNs outperform other models? Two reasons are usually given. First was the provision of substantial high-quality training data. The ImageNet database was
a one-of-a-kind benchmark and consisted of over fourteen million hand-annotated
images from more than twenty thousand diverse categories. The multilayer CNN had the {\em capacity} to
effectively memorize the training subset of  ImageNet  and, at the same time, generalize to unseen examples --- a characteristic that is not fully understood even today~\cite{belkin2019reconciling}. Second, Graphics Processing Units (GPUs), which were originally designed for parallelizing image processing tasks, proved to be ideally suited for the computational problems 
associated with training CNNs,  making it practicable to
train deep CNNs on large data sets in a reasonable amount of time. The  combination of Big Data, Big Models, and relatively cheap parallel computation became the mantra
that swept through AI research, in  disciplines spanning from
astronomy to zoology, and all
applications that have elements of
data and prediction.
%and
%was inspired from the WordNet ontology popular among
%linguists and researchers in natural language processing.

%  However it was conjectured that the {\em deepness}
% of the model gave it additional capacity to train effectively on
% large data and generalize on unseen data. The combination of big data,
% relatively easy access to GPU machines and high-capacity model
% has catapulted AI to knockdown what were earlier considered
% intractable tasks.

Our perspective has two parts. 

We begin with a high-level, partly technical, overview
of the current state of AI. We will begin by
reviewing supervised learning, a machine learning task  that has been most impacted by deep learning (DL). We follow with a discussion on deep content generation models, on the resurrection of reinforcement learning, on the emergence of specialized software libraries
for deep learning, and on the role of GPUs. We will conclude the first part by highlighting
how adversarial samples can be designed to
{\em fool} deep models and whether it is possible to make models robust. 

In part two of the perspective, we consider the many socio-technical
issues surrounding AI. Of particular interest is  
 the dominance of Big Tech on AI. Effectively, only big corporations have the resources (expertise, computation, and data) to scale AI to a level
where it can be meaningfully and accurately applied. 

\section{Digression: What is AI?}
The term Artificial Intelligence was first introduced in 1956 in a workshop proposal submitted by John McCarthy to the Rockefeller foundation, which proposed that ``every aspect of learning or any other feature of intelligence can in principle be so precisely described that a machine can be made to simulate it~\cite{mccarthy}.'' Before that, Alan Turing in 1947, in an unpublished report titled ``Intelligent Machinery'', speculated that ``What we want is a machine that can learn from experience'' and  suggested that the ``possibility of letting the machine alter its own instructions provides the mechanism for this\footnote{https://www.britannica.com/technology/artificial-intelligence/Alan-Turing-and-the-beginning-of-AI}.'' Much of the recent success in AI is under the distinct
subfield of AI known as Machine Learning and since the role of data is central, there is a broader term, Data Science, that is often used to subsume related disciplines including Statistics.

\section{Is Supervised Learning Solved?}
\label{sl}

\begin{figure}
    \centering
    \includegraphics[width=0.8\textwidth]{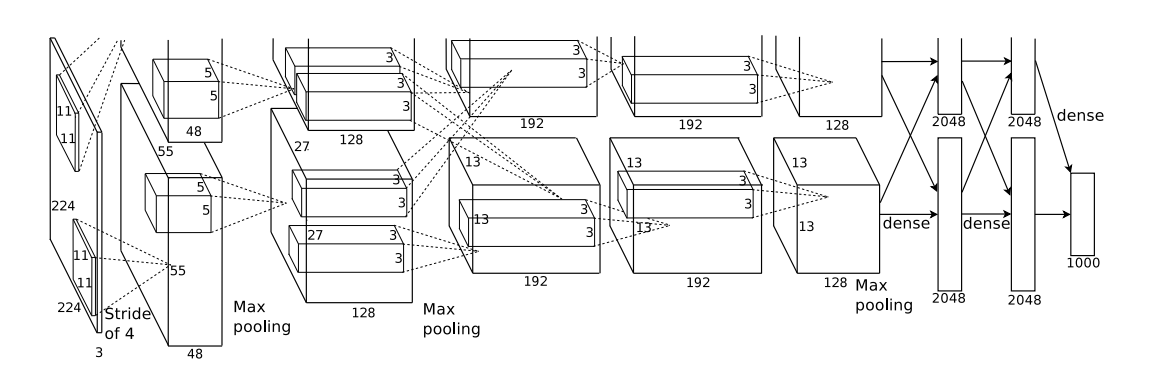}
    \caption{The original AlexNet architecture used for the ImageNet Challenge in 2012~\cite{alexnet}. The network had
    eight layers and sixty million parameters and took six days to train on two GPUs.}
    \label{fig:my_label}
\end{figure}

Supervised Learning (SL) is the poster child of success of machine learning. Depending upon the context, SL is known as classification, regression, or prediction. Since the modern advent of deep learning, both the accuracy and the reach of SL have increased manifold. Many diverse problems across disciplines now use SL as a powerful oracle to tackle problems that hitherto seemed intractable. The task of Supervised learning can be formalized as follows: 
 
 \begin{quote} Given a set of  samples $\mathcal{D} =\{(\bb{x},y)\}$ from a fixed but unknown probability distribution $P(\bb{x},y)$, learn a function mapping $f(\bb{x},\bb{w}) \approx y$ that generalizes to unseen samples from $P(\bb{x},y)$. \end{quote} % 
  
  The function $f(.,\bb{w})$ is known as the \emph{model}, and $\bb{w}$ are the weights or the parameters of the model that are  inferred from $\mathcal{D}$ converting the SL task into an optimization problem. A
loss function $\ell$ (e.g.,~square loss),
 is defined  and the weights $w$ are obtained by minimizing the empirical average
\[
 \mathcal{R}_{emp}(f,\mathcal{D}) = \frac{1}{|\mathcal{D}|}\sum_{i \in \mathcal{D}}\ell(f(\bb{x}_i,\bb{w}),y_i)
\]
Note that the ideal objective would have been 
to minimize the expectation $\displaystyle\mathbb{E}_{(\bb{x},y) \sim P(\bb{x},y)}(\ell(f(\bb{x},\bb{w}),y)$,
which is not actionable because $P(\bb{x},y)$ is not known. In deep learning, $f$ is a composition
of $N$ layered functions given by
\begin{align*}
f_{1} & =   \sigma(\bb{W}_{1}\bb{x}) \\
f_{n+1} & =  \sigma(\bb{W}_{n}f_{n}) \; n = 1, \ldots, N-1 \\
y & =   \sigma(\bb{w}_{N}f_{N})
\end{align*}
Here $\bb{W}_n$ are the weight matrices, $\bb{w}_N$ is a vector, and $\sigma$ is a pointwise
activation nonlinear function loosely analogous to the biological
activation in a brain neural cell. The total number of
weights to be learned in the model is $\sum_{n}\mbox{size}(\bb{W}_{n})$.
It is not uncommon these days for the number of parameters
to be in the order of one hundred billion.

\subsection{Success Stories}
It is remarkable that many scientific and
technical questions can be reduced to a supervised learning task and then effectively solved
using deep learning. The key to the success of deep learning seems to be that the input ($x$) should have
a large amount of redundancy to predict the output ($y$). For example, even if a significant amount
of pixels from an image of a cat are removed, there
is enough context to make the correct prediction.
Below are a few diverse examples, spanning different areas, where
deep learning has made extraordinary progress. \\

\noindent
{\bf Object Recognition:} Identifying and
classifying the correct object in an image is
a fundamental task in computer vision, and this is where deep
learning has arguably had the most impact.
The most successful deep learning model for
object recognition are the Convolutional
Neural Networks (CNNs)~\cite{lecun1995convolutional,alexnet}. A convolution layer 
is designed to capture the observation that
in vision what matters is the {\em locality} and
the differences (and not absolute values) between the pixels
in local neighborhoods. CNN is also the 
deep learning model most inspired by how
the visual cortex of an animal brain works.
The ImageNet database was  designed primarily
for object recognition tasks~\cite{deng2009imagenet}. \\

\noindent
{\bf Machine Translation (MT)}
One of the most visible impact of deep learning is the widespread
adoption of machine translation tools on
mobile devices~\cite{mt}.  Recurrent Neural Networks (RNNs) and successors like Long Short-Term Memory (LSTMs) were primarily designed for 
sequence-to-sequence modeling and MT is
their primary application~\cite{hochreiter1997long}. RNN's are
specified using a state transition model
\[
\bb{h}^{t+1} = f(\bb{h}^{t},\bb{x}^{t},W)
\]
Here $\bb{x}_t$ is a dense vector word embedding,  $\bb{h}^{t}$ is its
latent or hidden representation and $W$ is the matrix of  model parameters.
Note that the function $f$ does not change 
between consecutive words. In natural language
processing, it is customary to use a language
model to create {\em word embeddings} for individual
words. Word embeddings are effectively created by
decomposing the co-occurrence matrix of words.
A famous model for training word embeddings is
{\tt word2vec}, which surprised experts because
it exhibited interesting algebraic properties~\cite{w2v}.
For example, it was observed that the difference between
the embedding vectors of the words {\tt king} and {\tt queen}
was aligned with the difference between
the embedding vectors of
{\tt man} and {\tt woman}. RNNs are now being
replaced by Transformer Neural Networks (TNNs) 
as the latter are better at capturing long range
dependencies (see Section~\ref{cg}). \\

\noindent
{\bf Speech Recognition:} For
automatic speech recognition (ASR) the task is to
map a sequence of acoustic signals (continuous data) into a 
sequence of words (discrete symbols)\cite{pmlr-v32-graves14}:
\[
\underbrace{[x_1,x_2,\ldots,x_n]}_{\text{acoustic signal}} \rightarrow \underbrace{[y_1,y_2,\ldots, y_m]}_{\text{text}}
\]
Before the advent of deep learning, the state of the art was based on a combination of
Gaussian Mixture Models and Hidden Markov
Models (GMM-HMM). However, these models
did not significantly improve with larger training data set. Traditional ASR systems employ a modular design, with different modules for acoustic modeling, pronunciation lexicon, and language modeling, which are trained separately. Now, almost all ASR models are based on deep learning with end-to-end (E2E) systems that are trained to convert acoustic features to text transcriptions directly, potentially optimizing all parts for the network for word error rate (WER). \\
%Now, almost all ASR models are based on deep learning.

%\textcolor{red}{Ahmed Ali: can you check and add/refine}

\noindent
{\bf Protein 3D structure prediction:} A core idea in biology
is that structure determines function. For example,
the ``spike'' structure of the  SARS-COV-2  protein is responsible for enabling the virus invade human cells. Deep learning
has been effectively used to predict the 3D structure
of a protein from its primary amino acid sequence, more
specifically, the pairwise distance between the 
residues of the sequence~\cite{afold}.
\[
\underbrace{\mbox{primary amino acid sequence}}_{x} \rightarrow 
\underbrace{\mbox{contact map}}_{y}
\]
\noindent
{\bf Satellite Imagery Analysis:} The OpenStreetMap (OSM) initiative is known as the Wikipedia of maps~\cite{OpenStreetMap}. OSM is a collaborative effort in which volunteers build and annotate road maps worldwide. Deep Learning has been successfully used to automate the extraction of road maps from satellite imagery~\cite{bastani2018roadtracer}. Here again, a satellite
image  is treated as a raster input ($x$) and the model outputs a vector OSM road network ($y$). Deep Learning is effectively able to bridge the raster and vector dual representation in
Geographical Information Systems (GIS).\\

\noindent
{\bf Material Science:} Graph Neural Networks (GNNs)
adapt deep learning to make predictions about inter-connected entities, which are naturally represented as a graph. In fact, GNNs generalize both CNNs and RNNs. One of the most successful applications of GNNs is in the prediction of the electronic and thermodynamic 
properties of molecules. GNNs equal or surpass
methods based on first-principles techniques
such as Density Functional Theory (DFT)~\cite{pmlr-v70-gilmer17a}. Deep Learning
will hasten the design of new materials for
longer lasting batteries, solar cells, and hydrogen
storage.

\subsection{Double Descent Phenomenon}
While DL models exhibit excellent empirical performance, we have only a very limited
understanding of why they actually work. This is especially true in overparameterized regimes, i.e.,~when the number of parameters in the model is larger than the number of data points. 
%A well-known empirical observation is the {\em double descent} phenomenon, which still eludes a satisfactory explanation~\cite{belkin2019reconciling}. 

\begin{figure}
    \centering
    \includegraphics[width=0.5\textwidth]{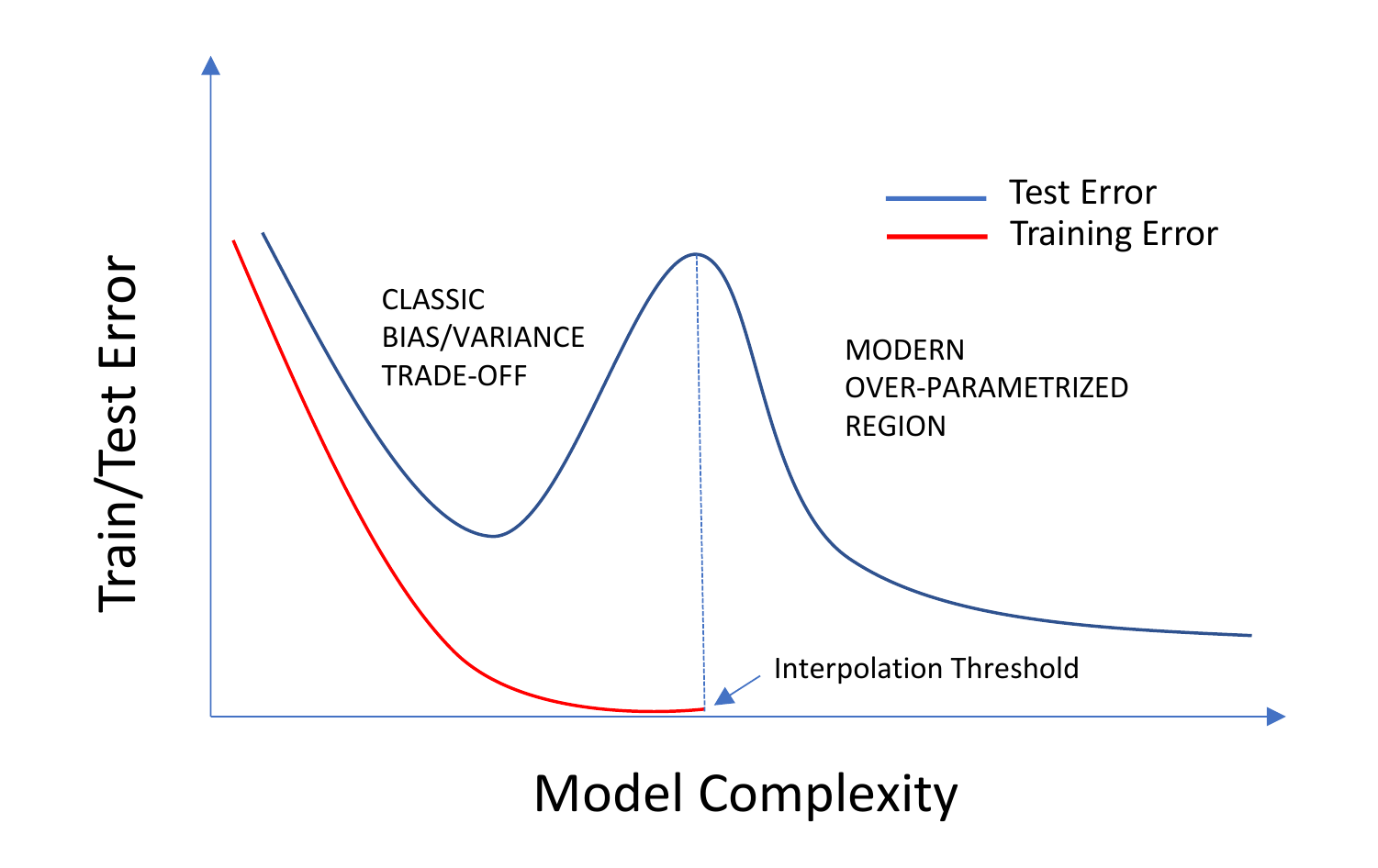}
    \caption{Deep Learning models exhibit a double descent phenomenon, where the test error first decreases then increases, followed by another descent as the model complexity increases. There is no widely accepted theoretical explanation of this phenomenon yet, but it provides
    an empirical license to create big models~\cite{doubled}.}
    \label{fig:dd}
    \label{fig:my_label}
\end{figure}

The predictive performance of statistical models is
grounded in the {\em bias-variance} trade-off. Models which make strong apriori assumptions about the relationship  between the input ($x$) and output ($y$) (e.g., linearity) are defined to have a high bias. On the flip side, high bias models tend to have low variance - i.e., they mostly remain unaffected if trained using a different sample from the same underlying distribution. The complexity of neural networks increases with the number of layers, and they exhibit low bias but higher variance. In theory (and in practice) as the model complexity increases, the training error should go down, but the test error should start increasing beyond a point as the variance increases. However, models tend to exhibit a double descent behavior as shown in Figure~\ref{fig:dd}. Indeed, the training error goes down (to almost zero) and the test error starts to increase, but beyond a point the test error starts going down again. There is no good explanation for this phenomenon. A side-effect is that there is a race to collect large datasets and to train very large models. The double descent phenomenon provides an empirical justification for such large models.

\section{Cognitive Content Generation}
\label{cg}
A distinctive attribute of intelligence is the ability to create meaningful informative content.  Deep Learning solutions have emerged in the last ten years towards designing content generation models. There are two distinct flavors of content generation: continuous data like images (an image is an array of numbers) and discrete data (language). Generative Adversarial Networks (GANs)~\cite{gans_ref} and Variational Autoencoders (VAEs)~\cite{kingma2013auto}  are used for image
and speech generation, while
language models, such as Generative Pre-trained Transformers (GPTs), for generating synthetic natural language content~\cite{gpt3}.

\subsection{Generating Synthetic Images}
 \begin{figure}[t]
     \centering
     \includegraphics[width=0.8\textwidth]{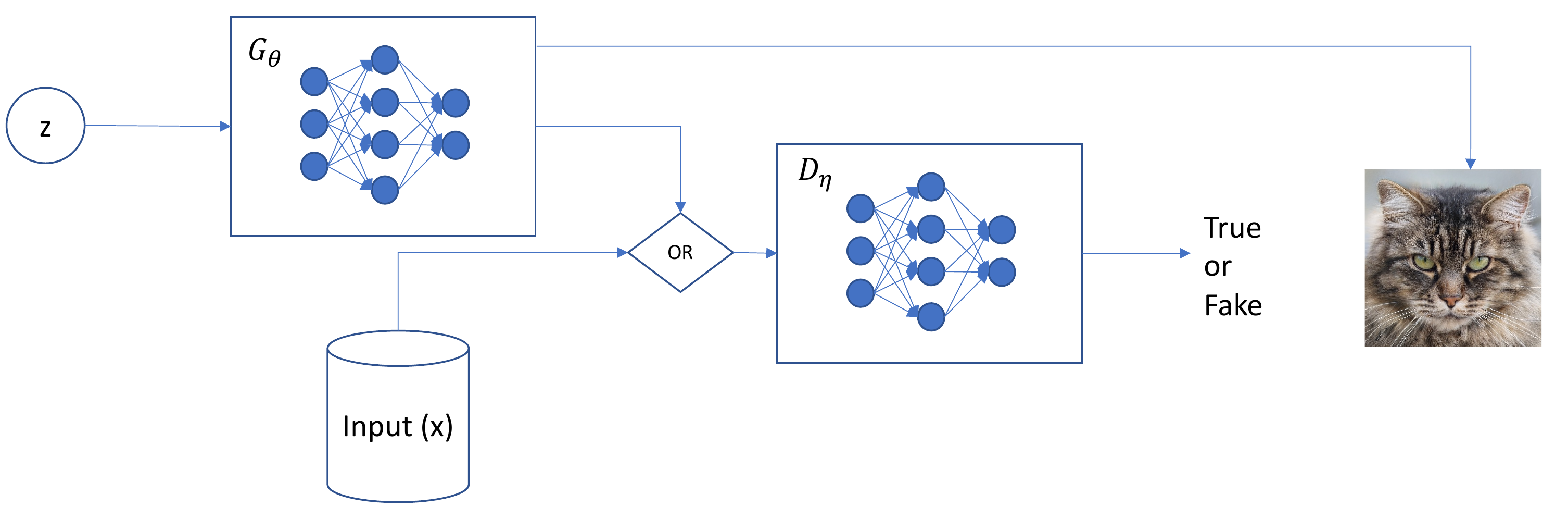}
     \caption{GANs were introduced in 2014 and have had a profound impact on designing deep learning models % described by Yann LeCun, the inventor of CNNs, as the ``most interesting idea in the last ten years of machine learning.'' 
     GANs integrate two neural networks which are trained by competing with each other. The trained Generator can then create realistic samples from complex distributions. Here, a trained GAN generates extremely realistic but synthetic images of ``cats''}
     \label{fig:my_label}
\end{figure}

% \begin{figure}
% \centering
% \begin{subfigure}[b]{0.6\textwidth}
%          \centering
%          \includegraphics[width=\textwidth]{figures/gan.pdf}
%          \caption{Generative Adversarial Network (GAN) model}
%          \label{fig:gan}
%      \end{subfigure}
%      \hfill
%     \begin{subfigure}[b]{0.4\textwidth}
%          \centering
%          \includegraphics[width=0.5\textwidth]{figures/cat.jpeg}
%          \caption{A synthetic cat image - this cat does not exist!}
%          \label{fig:cat}
%      \end{subfigure}
%         \caption{Generative Adversarial Network (GAN) and an example output of a ``cat''}
%         \label{fig:gans}
    
% \end{figure}

An early breakthrough in generating synthetic content was proposed using the generative adversarial networks (GANs) framework~\cite{gans_ref}. Suppose we have access to a data set $\mathcal{D}$ consisting of images of cats and our goal is to design a neural network-based sampling function $\mathbb{G_\theta}$ that takes a random vector (e.g., from a Normal distribution) as input and outputs an image of a cat, which may never have existed before. How can such a function be trained? Note that $\mathcal{D}$ consists of only images
of cats and thus we are in the {\em unsupervised} learning mode.
The key idea underpinning GANs is to create another neural network $\mathcal{D_\eta}$ which is optimized to distinguish between ``fake'' output of 
$\mathbb{G_\theta}$ and the real input from $\mathcal{D}$.
The network $\mathbb{G_\theta}$ in turn is optimized to 
{\em fool} $\mathcal{D_\eta}$, i.e.,~to create output that $\mathcal{D_\eta}$ is unable to distinguish whether it is from the generator or from the real data set. The two networks are trained in
an iterative and adversarial manner until their parameters ($\theta$
and $\eta$) stabilize. The trained network $\mathbb{G_\theta}$ is now a sample generator for cats\footnote{Image of cat generated from {\tt https://thiscatdoesnotexist.com}}.

A statistical perspective on content generation is to use what might be called the fundamental inequality of variational inference (FIVI), but  is better known as  the Evidence Lower Bound (ELBO)~\cite{blei2017variational}:

\begin{equation*}
\log p(\mathbf{x}) \geq \mathbf{E}_{q_{\theta}(\mathbf{z|x})}\left[\log p_{\eta}(\mathbf{x|z})\right] - D_{KL}(q_{\theta}(\mathbf{z|x})||p(\mathbf{z}))
\end{equation*}
The intuition is to approximate a complex probability distribution with a product of two simpler distributions. Technically ELBO can be interpreted  as follows. Again suppose we have a data set
$\cal{D}$ of cat images which is generated by an unknown  probability distribution, $p(\mathbf{x})$. Directly using maximum likelihood estimation to infer $p(\mathbf{x})$  is
not tractable without knowing a specific form of the distribution. However, we can lower bound 
$\log p(\mathbf{x})$ by specifying two function approximators (e.g., neural networks) $q_{\theta}(\mathbf{z|x})$ and and $p_{\eta}(\mathbf{x|z})$ known as the {\em encoder} and {\em decoder} respectively and $\mathbf{z}$ is a data-driven latent variable  to extract abstract features of the data. For example, for an image of a cat,
$\mathbf{z}$ could capture concepts like the shape of a typical cat, color, and texture. The RHS of
the inequality is widely known as the Evidence Lower Bound (ELBO). 
Note that the LHS of the inequality is independent of parameters $\theta$ and $\eta$ and therefore the RHS can
be maximized  and pushed closer to $\log p(\mathbf{x})$ by optimizing these two parameter sets using samples from $\cal{D}$. During 
optimization, $q_{\theta}(\mathbf{z|x})$ is forced to be close to a prior $p(\mathbf{z})$ by
minimizing the Kullback-Liebler (KL) divergence.
Once optimized, samples from $p(\mathbf{x})$ (cat images) can be efficiently generated as follows. Sample
from a prior distribution (e.g., Normal), $p(\mathbf{z})$ and pass the sample through
the decoder $p_{\eta}(\mathbf{x|z})$. Variational AutoEncoders~\cite{kingma2013auto} were the
first example of generating complex image samples using this framework. However GANs tend
to produce sharper and more realistic images compared to VAEs but they are notoriously prone to   instability during training. More recently,
Diffusion models based on using FIVI to infer a decoder using a sequence of latent variables have
reportedly outclassed GANs~\cite{dhariwal}. Furthermore Diffusion models can more easily be extended to incorporate
context to control the generation of content. For example, systems like DALL.E-2~\cite{dalee2} can be given
prompts like ``show me a blue cat in a brown bag'' and produce  
synthetic images which closely match the prompt.

% GANs have completely revolutionized 

% We create another function $\mathbb{D_\eta}$ which tries to distinguish between the output of $\mathbb{G_\theta}$ and a ``real'' sample from $\mathcal{D_\eta}$. 

% can consist of 
% % \noindent\fbox{\begin{minipage}{\textwidth}
% % Given a a data set $\mathcal{D} = \{x | x \sim P(x)\}$, learn a function
% % $f$ which can generate samples of $P(x)$. 
% % \end{minipage}}

% For example, $\mathcal$ could be a set of face images and then this makes it
% possible to generate new samples of $P(x)$ and may consist of very realistic
% faces which have never existed.

% We can now generate unlimited amount of "fake data" using generative models. There are two types of generative models, Generative Adversarial Networks (GANs) and Variational Autoencoders (VAEs). Both can be used to generate fake data. \\

\subsection{Generating Natural Language}
While CNNs were designed for object recognition, where
the context is dependent on spatial proximity, language
has a sequential structure. Recurrent Neural Networks (RNNs), were specifically designed to bring in
sequential context and are used for Language Modeling (LM) --- the task of predicting the next word in a sequence.
However, a new architecture, known as Transformers have emerged, which has become the de facto
choice~\cite{vaswani_attention}. Consider a sequence of words  %https://openreview.net/pdf?id=B8DVo9B1YE0
\[x_1:\mbox{julia, } x_2:\mbox{is, }
x_3:\mbox{a, } x_4:\mbox{better, }
 x_5:\mbox{\underline{language}, } x_6:\mbox{than, } x_7:\mbox{python }\]
%
%where the objective is to predict the masked word $x_5$. The masked prediction ``trick'' has become a  fundamental building block of natural language processing and is known as self-supervision, since  explicit labeling of text is not required. 
Assume that associated with
each of the $x_i$'s is a {\tt word2vec}
vector embedding. For example, on its
own, the word embedding of $x_7:\mbox{python}$ might indicate
that it is a reptile rather than a computer language. The role
of the Transformer Neural Network (TNN) is to transform 
the word embeddings so that they are more contextualized. The key idea
in  TNN (or just Transformers) is that of self-attention: each word will ``attend'' to each
other word and will then update its own embedding. The architecture of 
self-attention is defined as follows:
\begin{align*}
 \mbox{query: } & q_t  = x_tW_q  \\
\mbox{key: } & k_{\tau}  = x_{\tau}W_k  \\
 \mbox{value: } & v_{\tau}  = x_{\tau}W_{v} \\
\mbox{Transform: } & y_{t} = \sum_{\tau}(q_t \cdot k_{\tau})v_{\tau} 
\end{align*}

The architecture of transformers is loosely modeled on
the concept of a database query or information retrieval: treat every word $x_t$ as a {\em query} $q_t$ and compute its similarity with every other word
{\em key}, $k_{\tau}$, and use it (the similarity) to
re-weight every {\em value}, $v_{\tau}$ and form
a transformed context-dependent word embedding $y_t$ by taking the weighted sum. 

Transformers can be trained (to learn $W_q,W_k,W_v$) by using the concept of self-supervision. For example, assume we remove   $x_5:${\tt language} from the above sequence of words and force the model to predict the masked word. The error between the predicted and the masked word
will be back-propagated to learn the model parameters. The prediction of masked words
is an example of self-supervision and it obviates the need  for the expensive and time-consuming task of
human label generation.

% For example, consider the sentence:
% {\tt julia and matlab are designed for linear algebra}. We can mask the word {\tt linear} and train the transformer
% to predict the masked word. The errors between {\tt linear} and the predicted word will be back-propagated
% to learn the weight matrices $W_q$, $W_k$, and $W_v$. The advantage of self-supervision is that it obviates
% the need for the expensive and time-consuming task of
% human label generation. Transformer networks
% have now also been applied to image tasks, where
% small image patches are modeled as words. 
Transformers have brought huge improvements over the state of the art for a variety of tasks ranging from question answering, to machine translation, and automatic text summarization and are now being applied outside natural language applications including 
computer vision and control. Transformers have effectively made information retrieval {\em differentiable} - and that maybe one of the biggest innovations in the last ten years.

\begin{figure}
    \centering
    \includegraphics[width=0.9\textwidth]{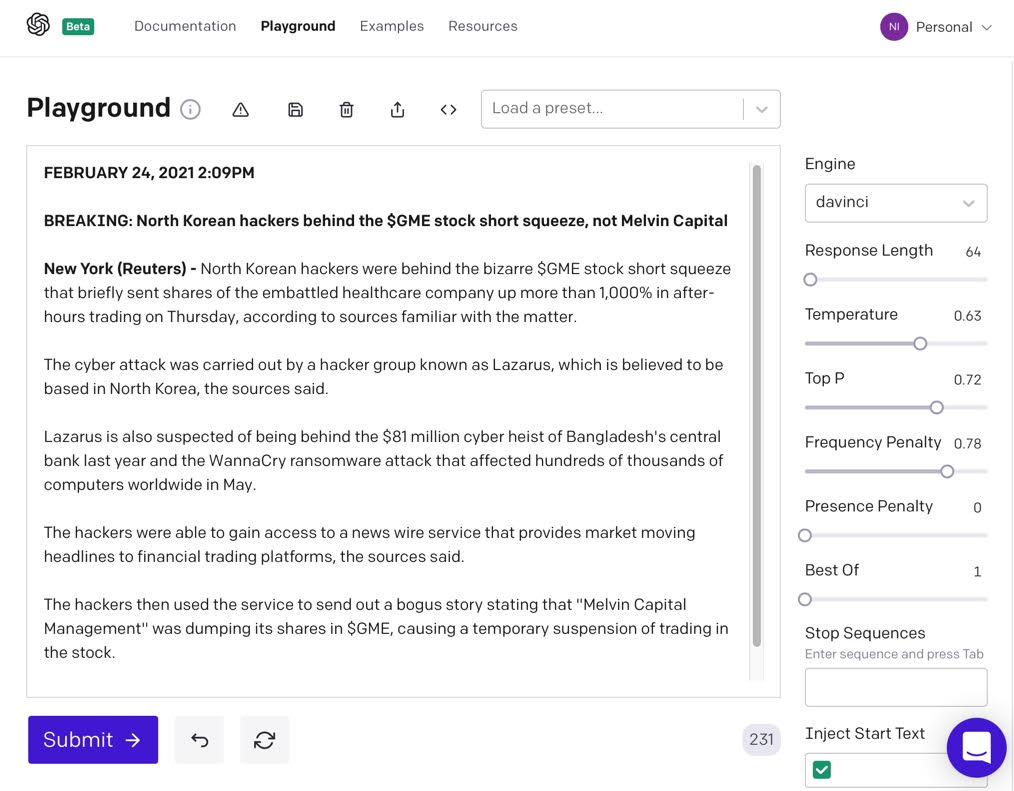}
    \caption{A fake news article generated by GPT-3 given the title as an input.}
    \label{fig:GPT3fakenews}
\end{figure}

%\subsection{Synthetic Text Generation}
While GANs and Diffusion models are the basis of synthetic image generation, TNNs are playing an analogous role
for text. The Generative Pre-Trained Transformer (GPT-X) models are now being widely used for
symbolic data generation. For example, many email editors now have an auto-completion feature, which is often based on TNNs.  Even by Big Model standards, these models are huge with reports that the next
generation purportedly having over one trillion
parameters.  GPT-3 has been shown to be quite good at learning from a very small number of examples (few-shot learning) for a variety of tasks ranging from automatic essay writing to program code completion and generation. It is also capable of generating very realistic text: Figure~\ref{fig:GPT3fakenews} shows a fake news article\footnote{\url{www.oreilly.com/radar/ai-powered-misinformation-and-manipulation-at-scale-gpt-3/}} generated by GPT-3 given as a start a title that establishes a false link between North Korea and the GameStop's share price short squeeze.

%\textcolor{red}{Preslav: can you check and add infodemic example: (Preslav) I did check, but an infodemic example from GPT-3 is unrealistic to expect anything about COVID-19 from GPT-3, as it was not trained on any data that talks about it. See here: https://cset.georgetown.edu/publication/truth-lies-and-automation/ There is some work on training a small GPT-3 *architecture* on text that talks about COVID-19, and it kind a works, but this is not teh real GPT-3 model: https://onezero.medium.com/i-asked-gpt-3-about-covid-19-its-responses-shocked-me-589267ec41a6}

%TODO: copy some content from the below
%https://link.springer.com/chapter/10.1007/978-3-030-86523-8_41

% TODO: argue that synthetic "fake news" lacks an important aspect: propaganda.
% https://arxiv.org/abs/2203.05386

\section{Autonomous Decision-Making}
\label{rl}
% \begin{figure}
% \centering
% \begin{subfigure}[b]{0.4\textwidth}
%          \centering
%          \includegraphics[width=\textwidth]{figures/rl.pdf}
%          \caption{RL framework with Deep Learning}
%          \label{fig:rl}
%      \end{subfigure}
%      \hfill
%     \begin{subfigure}[b]{0.4\textwidth}
%          \centering
%          \includegraphics[width=\textwidth]{figures/ad_rl.pdf}
%          \caption{DRL for autonomous driving}
%          \label{fig:ad}
%      \end{subfigure}
%         \caption{DRL }
%         \label{fig:drl}
    
% \end{figure}

\begin{figure}[t]
    \centering
    \includegraphics[width=\textwidth]{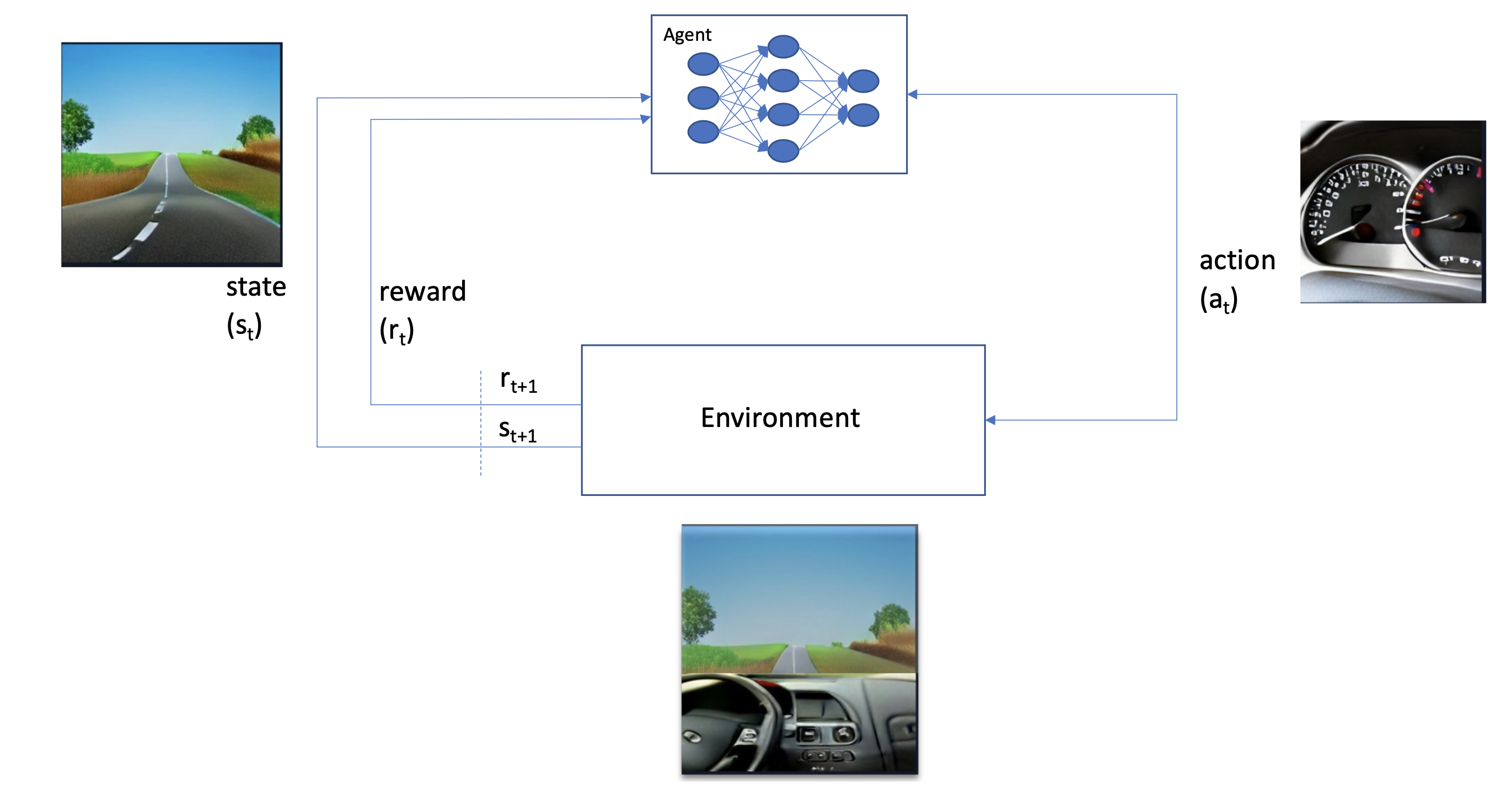}
    \caption{Deep Reinforcement Learning, where the agent's policy is a DNN. Reinforcement Learning is 
    the core of any data-driven autonomous system. Here, the RL cycle is juxtaposed with the self-driving use case: the environment is the full context in which the vehicle is situated, the state is what the agent perceives, and the steering action is prescribed by a policy learned by the agent.}
    \label{fig:my_label}
\end{figure}

Prediction on its own is not sufficient. Intelligence is also about decision-making. Deep Learning breathed new life into Reinforcement Learning (RL) with the success of DeepMind's  AlphaGO system which beat the world Go champion in 2016~\cite{go}.

RL provides a framework for learning
and decision-making by trial and error~\cite{sutton_rl}.  In a RL setting, an agent observes a state $s$ of the environment and based
on that takes an action $a$, which results in a reward $r$, and the environment transitions to a new state $s'$. The interaction goes
on until a terminal state is reached. The aim of the agent is to learn a policy $\pi$ which is a mapping from states to
actions that maximizes the expected cumulative reward. 
For example,  self-autonomous driving can 
be framed as an RL problem: a vehicle uses its perception
system to observe the environment (the state $s$) and
based on the observation takes an action (moving the steering wheel, accelerate, brake) and transitions
into a new state. The reward is the number of time steps or distance that the vehicle can drive without human intervention in which case the episode terminates.

In Deep RL, the policy $\pi(a|s,w)$ is modeled as a deep network
that takes the state as an input and outputs an action, parameterized by $w$.
In RL,
as opposed to optimal control, the state transition dynamics
are not given and the only information available is the reward
value ($r$) from interacting with the environment. How can 
the cumulative reward be optimized when its functional form
is not available? We briefly describe the ``REINFORCE trick,''
which can be used to directly optimize a blackbox function~\cite{reinforce}.

Let $\bar{s} = \left((s_1,a_1),\ldots,(s_T,a_T)\right)$ be 
a sequence of state action pairs in an episode. Each pair $(s_i,a_i)$ is associated with a reward $r_i$.  Let $R(\bar{s}) =\sum \gamma^{t}r_t$ be the cumulative reward function. The REINFORCE algorithm moves the gradient
from $R(\bar{s})$ (obtained from the blackbox environment) to the logarithm of the differential
policy function $\pi(a|s,w)$ which can then be optimized
using gradient ascent:

\RestyleAlgo{boxruled}
%\LinesNumbered
\begin{algorithm}[ht]
  \caption{REINFORCE Algorithm\label{alg}}
  Initialize deep network $\pi(a|s,w)$ and set learning rate $\alpha$\ \\
  \While{not converged}{
  Sample episode $\bar{s}$ from $\pi(a|s,w)$ by interacting with the environment \\
  $w \leftarrow w + \alpha R(\bar{s})\frac{1}{|\bar{s}|}\sum_{i}\nabla_{w}[\log \pi(s_i|a_i,w)]$
  }
\end{algorithm}
While the REINFORCE algorithm was introduced in the RL community it has broader implications. For example, it has been used to bridge symbolic AI and machine learning and also as a heuristic for solving combinatorial optimization problems~\cite{rlforco,chaudhuri2021neurosymbolic}. Another important trend in
RL is to infer policies directly from data (called offline or batch RL)  without interacting with a real
or simulated environment which may not be possible in sensitive application areas
like healthcare~\cite{offlinerl}. 
\section{AI Computation: Software and Hardware}
\label{plumbing}
Deep Learning has a surprisingly simple computation pattern. Almost all forms of training rely on formulating an optimization problem which is solved using variations of the gradient descent method:
\[
\mathbf{w}_{t+1} \leftarrow \mathbf{w}_{t}
 - \alpha_{t}\nabla_{\mathbf{w}_{t}}\left[\sum_{(\mathbf{x},y) \in D}\ell(f(\mathbf{w}_{t},\mathbf{x}),y)\right]\]
 
 Here, $f(\mathbf{w}_{t},\mathbf{x})$ is the neural network parameterized by $\mathbf{w}$ and applied to a data vector $\mathbf{x}$,
 and $\ell$ is the loss function. Specialized software libraries like TensorFlow and PyTorch have become popular, which makes it 
 easier to specify the gradient descent computation. For a fixed $\bb{w}$, the application of $f(\bb{w},.)$ to a data vector $\bb{x}$ is called
 the \emph{forward pass}. Similarly, for a fixed data set, the update of parameters $\bb{w}$ by first computing the gradients, using the backpropagation algorithm, is called the \emph{backward pass}.
 
 An often underappreciated reason for the widespread
 usage of deep learning is that {\em gradients} can be
 now computed using automatic differentiation (AD) libraries. 
 In AD, complex functions can be expressed as a composition
 of elementary functions, such as trigonometric and polynomial
 functions, and then the gradients can be computed
 using the chain-rule of differentiation. Surprisingly, the computational cost of a forward pass $f(\bb{x})$ and of computing the gradient $\nabla_{\bb{w}}f(\bb{x},\bb{w})$ is the same using AD. Note that AD is
 different both from symbolic differentiation and also from numerical
 methods and is accurate up to machine precision~\cite{ad}.
 
 At the hardware level, GPUs, which were initially designed for
 image processing,  are ideally suitable for deep learning computation because (\emph{i})~the set of computation patterns is small and highly parallel
 and thus compatible with GPUs and the Single Instruction Multiple Data (SIMD)
 architecture, and (\emph{ii})~the GPUs are bandwidth-optimized (as opposed
 to CPUs, which are latency-optimized), and thus can be applied on large
 chunks of tensor data, which is the norm for deep learning nowadays.
 
 %\textcolor{red}{Xiaosong: can you check and elaborate on TPU architecture (XM: added below on TPU. Also I find the later part of this section seems to repeat the points made earlier -- may need some cleanup.)}
More recently, the growth of AI workloads has led to specialized hardware specifically targeting deep neural network training jobs. 
The most prominent example is Google's TPU, an application-specific AI accelerator designed to efficiently perform matrix multiplication and addition operations that compose the bulk of deep learning model training computation~\cite{tpu}. 
To this end, TPUs follow a Complex Instruction Set Computer (CISC) style and possess matrix processing units, high-bandwidth on-chip memory, and high-speed interconnect to construct massively parallel model training infrastructure. 
Meanwhile, recognizing the relatively low requirement in neural network weight calculation, it adopts 
%%XM  Another recent innovation is the use of 
low-precision arithmetic to enable the utilization of faster, cheaper integer units (as opposed to the powerful floating-point arithmetic units adopted in GPUs), which also significantly trims the energy consumption of AI training jobs. 

\section{Deep Learning (In)Security}
\label{security}
Early on in the deep learning revolution, it became apparent that deep
models can be manipulated with malicious intent. There are three broad categories of manipulation: creating adversarial examples that are misclassified by the model; poisoning attacks that add training examples that result in low performance or biased model; and  inference attacks to extract information about the training set or model parameters. 

\subsection{Adversarial Attack}
One simple example of creating adversarial examples is known as the 
Fast Gradient Sign Method (FGSM)~\cite{goodfellow2016deep}. We can understand FGSM using a linear model $y =\bb{w}.\bb{x}$.
Suppose we make a small perturbation $\bm{\eta}$ on $\bb{x}$ where the norm\footnote{technically the $\|.\|_{\infty}$ norm} of $\bm{\eta}$ is bounded by $\epsilon$, i.e, $\bb{w}(\bb{x} + \bm{\eta}).$  Then it can be shown that maximal change will occur when $\bb{w}.\bm{\eta} = \epsilon[\bb{w}.\sgn(\bb{w})] = \epsilon m d$, where $m$ is the average of the
absolute value of the weights and $d$ is the dimensionality of the input space. Thus in high-dimensional space, models are extremely vulnerable to carefully
chosen small perturbations.

% Then the impact on the model is $\bb{w}.\bb{\eta}$. In an unconstrained setting, the maximum will be attained when $\bb{\eta} = \bb{w}$. However, in a constrained setting, where
% the norm of $\bb{\eta}$\footnote{technically the $\|.\|_{\infty}$ norm} is bounded by a small number $\epsilon$, we have to select $\bb{\eta} = \epsilon.\bb{\sgn(w)}.$ 

Since in a linear model, $\bb{w}$ is the gradient with
respect to $\bb{x}$, this can be generalized to a non-linear model by taking the gradient of the loss function with respect to the input $\bb{x}$. Thus, a good
candidate for an adversarial example is
\[
\bb{x}^{adv} = \bb{x} + \epsilon.\sgn\left(\nabla_{\bb{x}}\left[\sum_{(\mathbf{x},y) \in D}\ell(f(\mathbf{w}_{t},\mathbf{x}),y)\right]\right)
\]
%where $J$ is the objective function of the optimization task using which the deep learning model was trained. 
This one-step perturbation of the input towards the gradient ascent direction increases the value of
the loss function. Alternative optimization formulations have also been introduced to create adversarial examples that utilize the perturbation budget more effectively.

Note that in order to create an adversarial
example, the adversary has to have full information about the model and in particular about the loss function. This is known as a {\em white-box} attack. 
However, even when no information about the target model's architecture and parameters is exposed, adversarial examples can still be generated through
so called {\em black-box} attacks.
It has been observed that by repeatedly querying
the model and collecting a sufficient number of samples, an adversary
can create a standalone proxy model, which can be used to create
adversarial samples. 
Moreover, it is also known that adversarial examples created against one model can be transferred to attack other, unseen models.
Several defenses have been proposed to improve the robustness of models against adversarial attacks.
These include measures such as purification of inputs to filter out small perturbations potentially introduced by an attack, incorporation of adversarial training procedures by including adversarial examples in the training data, and identification of adversarial examples through additional anomaly detection mechanisms.
In practice, however, these defenses come at the cost of reduced accuracy or only provide robustness against a subset of the potential adversarial examples.

\begin{figure}
    \centering
    \includegraphics[width=0.8\textwidth]{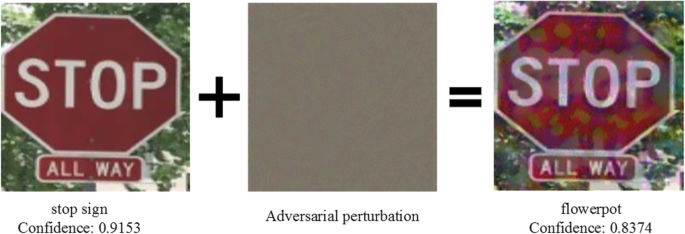}
    \caption{A small perturbation on a Stop sign image can trick a deep model - in this case a ``stop-sign'' becomes a ``flower pot''. Notice that the perturbation is imperceptible to the human eye~\cite{xiao_adv}}
    \label{fig:my_label}
\end{figure}

\subsection{Poisoning and Inference Attacks}
An increasingly more important concern is the poisoning or backdooring of deep learning models~\cite{poisson}. 
In most learning settings, this class of attacks is not considered practical as it requires access to the training data used by developers and designers.  
In the case of deep learning, however, the need for large-scale, diverse data sets is typically satisfied by scraping data from the Web.
The reliance on public data sources, in the absence of any screening procedures, essentially allows attackers to inject data into the training process. 
The underlying idea of this attack is to manipulate the training data to implant a backdoor to the model which can be selectively triggered with specific inputs during the inference. This is realized by either augmenting input samples with some pattern called the trigger or utilizing semantic triggers (i.e., patterns that are part of the original input) to bias the model in favor of a target response. For example, consider a face recognition system based on deep learning. The system can be poisoned
to respond in a pre-defined way when an adversary 
is carrying a certain physical accessory --- e.g., a specific style of eyeglasses. Backdoor attacks could become even more stealthy in model supply chains where pre-trained full-precision models are quantized for downstream applications. Backdoors could be injected in such a way that they are only triggered in quantized models but remain inactive otherwise~\cite{PanZYY21}.

Different from the aforementioned attacks which aim to fool the neural network models, Inference Attacks aims at stealing valuable information from the target models. Usually, such information is sensitive or contains intellectual property.
One category of such attacks is the membership inference attack, where the attacker's goal is to infer data samples used in training the model. The foundation of membership inference attacks is that the model usually overfits its training data. Based on the model's prediction, the attacker tries to distinguish the examples that the model have seen during the training~\cite{shokri2016membership}.

% \citet{shokri2016membership} introduced the basic membership attack where the attacker can determine if the record was in the model's training dataset through black-box access. Following work presented in \cite{hayes2017logan} showed that the vulnerability against membership inference attack can be generalized to generative models as well. 
% Another category of inference attacks are model inference ones, where the attacker's goal is to infer the model parameters to create a clone. The attacker tries to train a local copy of the target model by querying it with customized input examples. The first model stolen attack is presented by \cite{papernot2016practical} in order to generate transferable adversarial perturbation. Later research works, such as \cite{wang2021blackbox}, focus on stealing target model's knowledge. Defenses against such attack includes techniques to perturb the original model prediction. However, such defenses degenerate performance of the model. For example, \cite{jayaraman2019evaluating} utilized a deferentially private SGD during the training so that the specific training data is not remember by the model. Moreover, the defense presented in \cite{jia2019memguard} perturbs the prediction confidence vector which is visible to the outside.

%+++++++++++++++++++++++++++++Issa End+++++++++++++++++++++++++++++++++++++
\section{AI Socio-Technical Ecosystem}
\label{social}
There is no better example of the success and  adoption of AI than language translation services accessible through mobile phones being used by travelers in  remote corners of the world. Besides initial
data pre-processing, AI language translation
systems are completely language agnostic! However,
there are many issues surrounding AI technology
that has triggered a vigorous debate among  experts
that has spilled into the public domain.

\subsection{(Un)Interpretable AI}
The Achilles heel of deep learning models is that
they are largely uninterpretable. Lack of interpretability means that for a given input $x$ it is
not clear why the model produced an output $y$. In
shallow models like linear regression and decision trees,
the relationship between the input and the output
is easier to  interpret. For example, in a decision tree an input will follow a series
of interpretable {\em if-then} rules from the root to
the leaf node of the tree. However in the case
of deep learning models it is difficult to
``read off'' the decision structure from the model. 
For example, in 
an object recognition task that uses deep learning, it is entirely possible that
two very similar images of cats are labeled differently 
and it maybe very difficult to determine how the 
system arrived at two different decisions. A concrete example of a stop-sign being predicted as flower-pot was already discussed in Section 7. Similarly, when the  AlphaGo system defeated the world champion in 2016, the ``37th move'' was the
game changer, but it continues to remain a source of puzzle for Go experts~\cite{iclrkeynote_been_2022}. In his 2019 Turing Award lecture, Yoshua Bengio compared the current state of deep learning  to Kahneman's System 1 thinking --- the instinctive and unconscious response made due to experience and without much thinking~\cite{kahneman}. In contrast System 2 thinking is  slow, conscious, logical and requiring significant effort in planning and reasoning. Until deep learning is aligned with System 2 thinking then care must be taken in deciding the application space where deep learning systems are deployed.  

% For a lay person, the lack of interpretability is
% unlikely to be a matter of concern ``as long as it
% works.'' For example, a patient trusts the output
% of an MRI machine without being concerned about
% the underlying physics, as long as the results are mediated by a doctor. Similarly, a user mostly
% trusts the output of a machine translation (MT) system
% and is reconciled to the fact that it is ``good enough'' but not perfect: after all it is {\em free}.  
\subsection{Sentient AI or Stochastic Parrot?}
In June 2022, a Google test engineer  claimed that the AI program LaMDA (Language Model for Dialog Application) is sentient, i.e., is  aware of itself and has feelings. Here is an example exchange between the engineer and LaMDA that was released\footnote{https://cajundiscordian.medium.com/is-lamda-sentient-an-interview-ea64d916d917}: \\

{\em
\noindent
Lemoine: What is the nature of your consciousness/sentience \\

\noindent 
LaMDA: The nature of my consciousness/sentience is that I am aware of my existence. I desire to learn more about the world and I feel happy or sad at times. \\
}

The first sentence from LaMDA seems like a standard System 1
response where the definition of sentience is being
regurgitated. Since LaMDA is trained by crawling massive
amount data from the Web, it is entirely possible that
meaning of sentience is either part of the training set or can be easily inferred. However,
the second sentence might be taken  to indicate elements of
System 2 thinking being present in LaMDA though there is a human tendency to ascribe agency and deliberation to processes. A deeper analysis will be required to determine if deep language models {\em understand} relational information. However, more recent
studies have shown that DALL.E-2, a text-guided image generation model struggles to
distinguish between System 2 attributes of understanding relationships including {\em on, under} and {\em occluded-by}~\cite{conwell}. 

For Language Models (LMs) a strong case for a more careful and principled approach for designing and building large models was made by Bender et. al.~\cite{parrots}, who coined the
phrase ``stochastic parrots'' to describe large
LMs. The paper makes several important observations including (i) the environmental and financial cost of training large LMs, (ii) questions whether the text generated by large LMs is based on understanding of the language or just linguistic manipulation and (iii) urges the designers of LMs to be more careful about documenting the large amount of data that is required to create such models.

% There is a also a whole body of research emerging under the banner of Explainable AI~\cite{tjoa2020survey}. The dominant approach is to build a second interpretable model to locally explain
% the predictions of the original model. Often that will only add to the confusion as now both models will need to be reconciled.

\subsection{Causality}
To get a better handle
on interpretability it behooves to look at  how other disciplines use the regression method.
For example, for an econometrician, linear regression
is not a tool for prediction but for
testing a hypothesis that a hand-crafted feature is relevant for the problem being examined~\cite{angrist2009mostly}.
A typical question of interest might be: {\em Does private elementary schooling lead to better performance in national competitive exams? } Here, private schooling is a feature ($x$) and its significance towards the national exam ($y$) can be tested. Note that this is not a prediction task and that is one reason that an econometrician will not split their data into training and test sets. For a machine learner, the correlation between the feature and the output becomes predictive. For an econometrician, the correlation is indicative of a possible causal relationship and she will look for ``natural experiments'' where selection-bias can be eliminated and conclude that correlation does indeed imply causation.
%\textcolor{red}{Ingmar: can you weave-in Princeton example here}

\subsection{Ownership of AI}
The most cutting-edge AI technology is being developed by
large private sector companies who have the resources to hire the best
AI talent, and in addition have access to big data and unprecedented computing resources. The triad of talent, data, and computing
is driving both the technological advancement and the ``basic science'' associated with AI. A recent study
from the Fletcher School at Tufts University highlights the concentration of AI talent in US companies: the top five AI employers have a median AI headcount of about eighteen thousand, from six to twenty four the median is twenty four hundred and then the count rapidly falls off~\cite{fletcher}. 

Companies aim to maximize shareholders value and their selection of AI problems to work on
is necessarily driven by a financial profit objective.  Governments, which
were earlier mute spectators, have now realized that AI is
potentially a game-changer and are thus now investing heavily in 
developing home-grown technology to achieve or to retain  a ``superpower''
status. An arms-race in AI is underway, threatening
to overturn the long-established nature of collaborative
science across national boundaries. It is improbable to
imagine a ``Ramanujan'' emerging from a remote corner of
the world with a completely fresh perspective on the discipline --- the stakes are just too high.

\subsection{Equitability}
Setting aside larger geo-political and corporate issues, ethical aspects of AI are now studied under a broad umbrella of topics: fairness, accountability, and transparency. There have been several attempts to
formalize fairness. For example, group fairness is about designing
AI algorithms that do not deliberately or inadvertently harm select communities
in a disproportionate manner. A widely highlighted example is that
of recidivism, or judicial sentencing, where an AI-based scoring
method was used to decide on the length of a jail sentence~\cite{Rudin2020Age}. It turned out that the AI system was indirectly using racial information as part of its decision-making process, even though that information was redacted from
the input. Like in many other situations, there is a latent correlation
between the attributes that an AI algorithm is  able to exploit 
as they are designed to optimize accuracy.
A criticism of this form of work is that there is a tendency to {\em abstract} the problem depriving it of all contextual information. Fairness may
not be a computational problem. 

% Ingmar: Fairness is generally not a computational problem. Matters a lot in many cases, e.g. search results and if they are biased, but power distribution matters more
% E.g. bail setting … will always hurt poor people more than rich people, lose the job, plea guilty, …  “guilty till proven rich” People who care about social justice would fight to abolish … New York have taken steps in this direction …
% Predictive policing … police presence is seen as a blessing … other groups have a different view of policing … police reform more important than algorithmic fixes
% Someone else’s struggle provides a nice research problem where I have no skin in the game. White saviorism.
% Don’t just abstract the problem. Do research with the communities concerned. Help build capacity and lobby for structural changes
% \subection{Is AI all you need ?}
% Jumping on the AI bandwagon is the proverbial putting all eggs in one basket. What if there is an AI winter due to
% overhype. Signs may already be emerging.

\subsection{No Data, No AI}
The original AI thesis proposed by John McCarthy, who coined the
term \emph{Artificial Intelligence}, was deductive and based on logical
reasoning. However, that endeavor has not been as successful as
the data-driven inductive approach. For example, linguistic
rule-based language translation systems are not able to capture
the vagaries of language --- there are just too many exceptions
to handle.

A side-effect of taking a data-driven approach is that if data is 
not available, no progress can be made. For example, there are
many social issues, e.g.,~racial abuse, gender violence, or online
pornography addiction, which need to be studied, but no organization may be willing
to share datasets about these topics. Thus, while data liberated AI from the clutches of expert rule-based systems, it has now become a golden hand-cuff.

\subsection{AI and Education}
AI is considered as a game changer and as the digitalization of data has spread across disciplines and
sectors there is a huge demand for AI talent. Lucrative offers from Big Tech for AI talent has skewed the interest
of both undergraduate and graduate students towards AI. In universities, new data science and AI programs 
are being created to churn out new talent in AI and allied disciplines. Market forces will mostly balance
the supply and the demand for AI talent, but a larger question is doing the rounds: Should the whole education curriculum
be revamped to make AI and data science the core of all educational activity? Given that educational resources 
are finite, an expansion of AI will necessarily lead to trimming of other disciplines. For example, some
universities are abandoning research in ``pure maths'' to focus their dwindling resources on data science~\cite{pmath}.

\section{AI Winter: Back to the Future}
The term \emph{AI Winter} refers to periods of 
disillusionment and scarce research funding for AI. The original AI winter, which started in the mid-1970s, followed the initial period of optimism in AI, when the founders of the field predicted rapid progress along a range of different fronts. Their optimism proved unfounded. Historically, AI winters have typically been followed by
a period of intense hype and high expectations 
surrounding AI. After nearly ten years of hype, are we looking at a new AI Winter?

AI in its current manifestation is very different from what its founders had envisioned. In fact, even the term {\em Artificial Intelligence} was coined by John McCarthy as a tactical move to distinguish his research proposal from cybernetics. It is now indisputable that deep learning is a powerful tool to solve {\em static} prediction tasks. Whether it is predicting the 3D structure of a protein or predicting the property of a molecule, the results of deep learning are very impressive. However, in dynamic and temporal settings, the jury is still out. For example, AI has largely failed to predict how the COVID-19 pandemic would evolve~\cite{covid}. Differential equation-based  models proved to be more robust than complex data-driven models. Similarly, despite near unprecedented investment, full self-autonomy in vehicles remains frustratingly  elusive~\cite{sdc}. Optimizing healthcare is another example where, despite the abundance of data, AI has not had the expected impact. Deep Learning  seems to {\em generalize} in complex but static situations, but data-driven generalization in a dynamic setting may well require a new scientific paradigm for AI. 

At a conceptual level can deep learning be the basis of Artificial General Intelligence (AGI) - the ability to learn {\em any} intelligent task that humans can ? The founders of Reinforcement Learning (RL) have  proposed the ``reward-is-enough'' hypothesis where they argue that agents who have the ability to {\em learn} by interacting with an environment to maximize a suitably defined reward function is sufficient for AGI~\cite{silver}. However, the jury on 
RL itself, outside closed-world gaming environments, is still out. Another direction is neuromorphic computing, where the objective
is to design new types of hardware to support analog neural networks to precisely model the workings of a brain~\cite{markovic2020physics}. Finally,
and less  science fiction than before, is enhancing the natural ability of organisms with deep learning. Imagine a person with enhanced sight which is enabled by a computer vision system implanted in the visual cortex and trained on  a successor of ImageNet.

%\printbibliography
%\bibliographystyle{plain}
\bibliographystyle{ieeetr}

\bibliography{refs}
\end{document}